\documentclass{article}

     \usepackage[preprint]{neurips_2020}

\usepackage[utf8]{inputenc} 
\usepackage[T1]{fontenc}    
\usepackage{hyperref}       
\usepackage{url}            
\usepackage{booktabs}       
\usepackage{amsfonts}       
\usepackage{nicefrac}       
\usepackage{microtype}      
\usepackage{graphicx}
\usepackage{amsmath}
\DeclareMathOperator*{\argmax}{arg\,max}

\title{Efficient Ridesharing Dispatch Using Multi-Agent Reinforcement Learning}

\author{%
  Oscar de Lima\thanks{indicates equal contribution} \\
  University of Michigan\\
  Ann Arbor, MI 48109 \\
  \texttt{oidelima@umich.edu} \\
   \And
   Hansal Shah* \\
   University of Michigan\\
  Ann Arbor, MI 48109 \\
  \texttt{shansal@umich.edu} \\
   \AND
   Ting-Sheng Chu* \\
    University of Michigan\\
  Ann Arbor, MI 48109 \\
  \texttt{timchu@umich.edu} \\
   \And
   Brian Fogelson* \\
    University of Michigan\\
  Ann Arbor, MI 48109 \\
  \texttt{bfogels@umich.edu} \\
}

\begin{document}

\maketitle

\begin{abstract}
  With the advent of ride-sharing services, there is a huge increase in the number of people who rely on them for various needs. Most of the earlier approaches tackling this issue required handcrafted functions for estimating travel times and passenger waiting times. Traditional Reinforcement Learning (RL) based methods attempting to solve the ridesharing problem are unable to accurately model the complex environment in which taxis operate. Prior Multi-Agent Deep RL based methods based on Independent DQN (IDQN) learn decentralized value functions prone to instability due to the concurrent learning and exploring of multiple agents. Our proposed method based on QMIX is able to achieve centralized training with decentralized execution. We show that our model performs better than the IDQN baseline on a fixed grid size and is able to generalize well to smaller or larger grid sizes. Also, our algorithm is able to outperform IDQN baseline in the scenario where we have a variable number of passengers and cars in each episode.Code for our paper is publicly available at \href{https://github.com/UMich-ML-Group/RL-Ridesharing}{https://github.com/UMich-ML-Group/RL-Ridesharing}.

\end{abstract}

\section{Introduction}

Ridesharing services are becoming widely popular with services like Uber and Lyft. A recent study \cite{yaraghi2017current} estimated that the ridesharing economy will increase from \$14 Billion in 2014 to \$335 Billion by 2025. With this increasing demand, dispatching taxis efficiently has become a widely researched problem. Solving this problem is important since it has a two-fold advantage: helping increase customer satisfaction with shorter waiting times and increasing revenue for taxi services by completing more rides. Early works aimed to improve ridesharing efficiency \cite{agatz2012optimization,zhu2016public} but require handcrafted cost functions for estimating travel times and user waiting times. Clearly, these approaches suffer from inaccuracies when used in the dynamic real-world environment in which taxis operate. This complexity of the real world can be modeled more accurately using Reinforcement Learning (RL) based approaches.

In recent years, RL has achieved great success in modelling and solving extremely complex problems \cite{silver2016mastering,silver2017mastering}. Naturally, there has been work in the past to use RL for efficient ridesharing \cite{godfrey2002adaptive,godfrey2002adaptive2,wei2017look}. However these traditional methods are unable to accurately model the complicated dynamics involved in a real world environment. With the advent of Deep Reinforcement learning (DRL) \cite{mnih2013playing,mnih2015human} it has been possible to achieve super-human performance levels on previously intractable problems. In light of such advances, we propose a Multi-agent DRL based approach to solve the Ridesharing problem. \cite{al2019deeppool,jindal2018optimizing} use Independent DQN (IDQN) \cite{tampuu2017multiagent} based approaches which learn decentralized value functions or policies. These methods are prone to instability due to the non-stationarity of the environment induced by simultaneously learning and exploring agents. We use the approach of QMIX adopted from \cite{rashid:icml18} to achieve centralized training and decentralized execution. This allows the network to be trained with a global value function and enables the choice of greedy actions for each agent corresponding to its individual value function.

In summary, we propose a multi-agent RL approach based on QMIX \cite{rashid:icml18}, which achieves better quantitative results than IDQN based methods. Additionally, we model the problem effectively using a First Come First Serve (FCFS) principle in which passengers requesting the ride first are given higher priority over others, which should reduce passenger waiting time. By selecting a car for the passenger with the highest priority at a given time, the size of the  action space is reduced. Through experiments conducted in this project, it is demonstrated that a network trained on a fixed grid sizes can generalize well on other smaller or larger grid sizes and is able to perform better than the baseline when the number of passengers and cars are varied between each episode.

\section{Related Work}

Multi-Agent RL (MARL) has been considered an important problem in RL with rich history of works \cite{yang2004multiagent,bu2008comprehensive}. Earlier methods of work consisted of mainly tabular based methods, but more recent work is moving towards deep learning based methods \cite{tampuu2017multiagent,peng2017multiagent,foerster2018counterfactual} which can tackle high-dimensional state and action spaces. One relevant approach to this work is that of IDQN \cite{tampuu2017multiagent} where two agents interact with each other only through the reward. Although these approaches achieve decentralization, they are prone to instability which arises due to the non-stationarity of the environment induced by simultaneously learning and exploring agents. Attempts have been made in the past to alleviate some of these issues. \cite{omidshafiei2017deep,foerster2017stabilising} address learning stability but still learn decentralized value functions, whereas \cite{guestrin2002multiagent,kok2006collaborative,peng2017multiagent} adopt centralized learning of joint actions but require substantial communication between agents during execution. QMIX \cite{rashid:icml18} provides centralized training with decentralized execution. QMIX employs a network that estimates joint action-values as a complex non-linear combination of per-agent values that condition only on local observations. We thus adopt a method based on QMIX for our solution. 

Ridesharing, the focus of this paper has also been researched extensively in the past. Traditional methods \cite{godfrey2002adaptive,godfrey2002adaptive2,wei2017look} are unable to model the complex dynamics of the real world accurately. Algorithms proposed in \cite{powell2011towards,zhang2014carpooling,qu2014cost} require routing recommendations to be provided to drivers to maximize company profits. Multi-agent IDQN based approaches \cite{jindal2018optimizing,al2019deeppool} suffer from the disadvantages mentioned earlier. The approach that we adopt, based on QMIX \cite{rashid:icml18}, achieves superior performance over other MARL methods. This allows our network to perform well in realistic scenarios with variable number of passengers and cars in each episode. Generalizing capabilities were also achieved with a model trained on a fixed grid size, which performed better than baseline methods when tested on a smaller or a larger grid size.

\section{Methodology}
\label{sec:method}
In this section we outline the environment developed for this project, the Greedy and IDQN baselines implemented, and our proposed method based on QMIX \cite{rashid:icml18}.

\subsection{Environment}
\label{sec:sim_env}
\begin{figure}
    \centering
    \includegraphics[width=7cm]{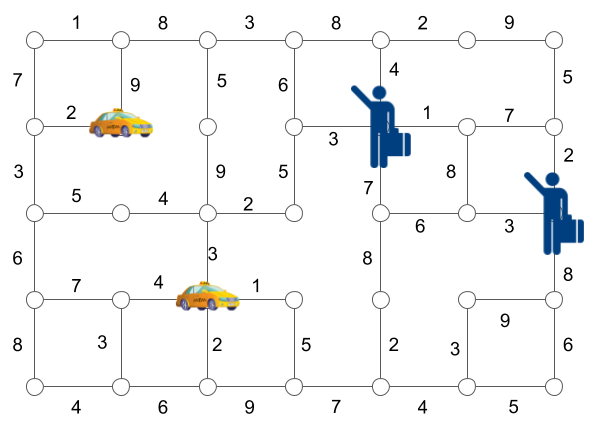}
    \caption{An example of simulation environment with randomly assigned road cost.}
    \label{fig:gridmap}
\end{figure}

In order to evaluate and compare the proposed algorithm, a simulated environment is implemented. To avoid a complicated implementation with a real-world map, a grid map is used. The nodes of the grid map correspond to different intersections while the edges between them correspond to different roads. Every road has a cost corresponding to the amount of time it takes a car to cross it. The costs are representative of factors including different traffic conditions. The costs are randomly assigned by the environment and cannot be observed by any algorithm. An illustration of the grid map is shown in Figure \ref{fig:gridmap}.


\subsection{Greedy Baseline}

A non-learning-based greedy algorithm is implemented as a baseline method and will be compared to various reinforcement learning methods. The greedy algorithm follows a first-come-first-serve (FCFS) strategy. As a result, passengers who requests the car earlier are given higher priority. Every passenger is paired to a car naively according to the distance. The Manhattan distance is used due to the nature of the grid map setting. No knowledge of the costs of the roads is used to make this decision. The performance of this greedy algorithm has been examined and the results are shown in Section \ref{sec:experiment}. The time to finish dropping off all passengers is then measured for multiple combinations of cars and passengers.

\subsection{Independent Deep Q-Learning (IDQN)}
\label{sec:dqn}

To dispatch cars to passengers in an efficient way, a reinforcement learning algorithm is introduced. Q-Learning has been proven to outperform previous state-of-the-art algorithms in scenarios similar to our problem, as shown in \cite{al2019deeppool,jindal2018optimizing}. Therefore, Q-Learning will be used as our baseline reinforcement learning algorithm. 

The problem is defined as follows. The maximum number of cars and passengers on the map is defined as $C_{max}$ and $P_{max}$. The state of each car on the map is defined as $s_c=(c_x, c_y)$ where $c_x$ and $c_y$ are the coordinates of the car. The state of each passenger on the map is defined as $s_p = (p_x, p_y, d_x, d_y)$ where $p_x$ and $p_y$ represent the coordinates of the pick up point, and $d_x$ and $d_y$ represent the coordinates of the drop off point for the passenger. Two binary indicator vectors,  $I_p$ of length $P_{max}$ and $I_c$ of length $C_{max}$, are used to tell the network which cars and passengers are in the environment. The observation of the environment is then $(s^1_c, ..., s^{C_{max}}_c,I_c, s^1_p, ..., s^{P_{max}}_p, I_p) \in \mathcal{S}$. If the $i$-th car or the $j$-th passenger do not show up on the map, then $c^i_x, c^i_y, p^j_x, p^j_y, d^j_x$ and $d^j_y$ are set to $0$. Each passenger is paired with a car as part of the overall action.

Deep Q-learning uses deep neural networks parameterized by $\phi$ to represent the action value function of an agent \cite{mnih2013playing}. To make the algorithm more computationally efficiency, the action value of every passenger for every action is computed at the same time by the same network. Therefore, the output of this network is a matrix whose rows represent the passengers and whose columns represents the car to which each passenger can be matched. The value of the matrix at (i, j) is the Q-value where the i-th passenger takes the action of pairing with the j-th car.
The dimension of the network output $\in \mathrm{R}^{\; P_{max} \times C_{max}}$. 

Each passenger is paired with the car which has the maximum Q-value in the row, following the FCFS order. Each passenger gets their own reward, which is (-1) $\times$ (the passenger's waiting time). By doing so, when the reward is maximized, the passenger's waiting time is minimized. An episode starts by randomizing the state of every car and passenger on the grid map. When we test the scenario with variable agents, we also randomize the number of cars and passengers between episodes. All passengers are paired with cars at the beginning of the episode and the episode ends when all passengers are dropped off. A step in this implementation is equivalent to an episode, so when we store a transition tuple $(s, a, r)$ in the replay memory we don't include the next state. The implication is that we also don't learn a target network. We use the Huber loss shown in eq. (\ref{eq:huber_loss}) \cite{huber1964} since it is robust to outliers when the estimates of the action-value are noisy and can prevent exploding gradients in some cases \cite{girshick2015fast}.

\begin{equation}
\label{eq:huber_loss}
H(x) = \begin{cases} 
      \frac{1}{2}x^2  &|x|\leq 1\\
      |x| - \frac{1}{2} & otherwise
   \end{cases}
\end{equation}

The loss function is then:

\begin{equation}
\label{eq:QMIX_loss}
L(\phi) = \frac{1}{B \times P_{max}} \sum_{i=1}^{B}\sum_{n=1}^{P_{max}} H({r}_{n}^{i} - Q_n(s, a_n, \phi))
\end{equation}

where $r_n$ is the reward for the $n^{th}$ passenger, $Q_n$ is the action-value for the $n^{th}$ passenger given the choice if action $a_n$, and $B$ is the batch size.





\subsection{QMIX: Monotonic Value Factorization}
\label{sec:QMIX}

While methods like IDQN and our greedy baseline try to maximize the reward of each agent independently, QMIX \cite{rashid:icml18} promotes coordination between agents by having a shared global reward $R_{tot}$ and learning a joint action-value function $Q_{tot}(s,a)$  where $s$ is the observation of the environment and $a$ is the joint-action of all the passengers. Consistency between the individual and global action-values is guaranteed by ensuring that the joint-action that maximizes the global action-value is equal to the set of actions that maximize the individual action-values as shown in equation (\ref{eq:consistency}).

\begin{equation}
\label{eq:consistency}
    \argmax_a Q_{tot}(s,a) = \begin{bmatrix}
  
  \argmax_{a_1} Q_1(s, a_1) \\
  \argmax_{a_2} Q_2(s, a_2) \\
  \vdots \\
  \argmax_{a_{P_{max}}} Q_{P_{max}}(s, a_{P_{max}})
         \end{bmatrix}
\end{equation}

Having a global state-action value allows every passenger to choose actions greedily with respect to their individual action-values $Q_a$. Also, if the condition in equation (\ref{eq:consistency1}) is satisfied, finding the $\argmax_a Q_{tot}$ becomes trivial.

We can ensure monotonicity with a constraint on the relationship between the global action-value $Q_{tot}$ and the action-value for each passenger $Q_p$:

\begin{equation}
\label{eq:consistency1}
\frac{\partial Q_{tot}}{\partial Q_{p}} \geq 0, \forall p \leq P_{max}
\end{equation}

\begin{figure}
    \centering
    \includegraphics[width=12cm, scale=.9]{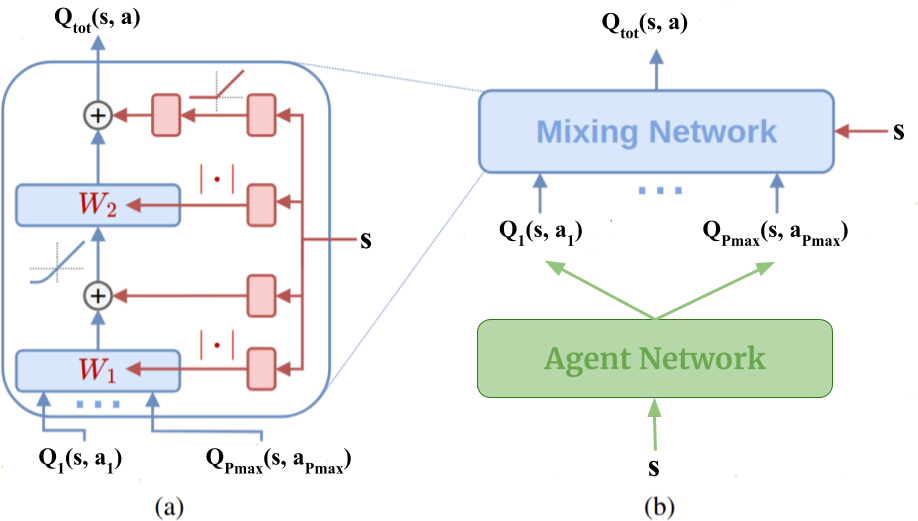}
    \caption{(a) Mixing network structure. In red are the hypernetworks that produce the weights and biases for mixing network layers shown
in blue. (b) The overall QMIX architecture. Adapted from Rashid et al. \cite{rashid:icml18}}
    \label{fig:QMIX}
\end{figure}

Our implementation has an agent network, a mixing network, and 2 hypernetworks \cite{ha2016hypernetworks}. The overall structure of the network can be seen in Figure \ref{fig:QMIX}. In the original QMIX implementation \cite{rashid:icml18}, each agent had their own agent network whose output was their value function $Q_p(s, a)$. For computational efficiency and simplicity we decided to use a shared network for all passengers like the one shown in section \ref{sec:dqn}.

When optimizing our model, the mixing network 
takes the action-values from the selected actions in the replay buffer as input and mixes them
monotonically to produce $Q_{tot}$. The
weights of the mixing network, but not the biases, are restricted to be non-negative. Dugas et al. \cite{dugas2009incorporating} showed that this allows any monotonic function to be approximated arbitrarily closely. 
We force the weights of the mixing network to be non-negative by using a separate hyperparameter network to produce them on every layer. The input to these hyperparatmeter networks is the state of the environment and the output is a vector which is reshaped to the appropriate size for that layer. We use a separate hyperparameter network for the weights of each layer and for each bias. The weights, but not the biases, are passed through an absolute activation function. The hyperparameter networks for the biases have one layer, except for the one used in the last layer of the mixing network which has 2 layers and a ReLU non-linearity. This architecture is shown in Figure. \ref{fig:QMIX}.a.

We train using the following loss:

\begin{equation}
\label{eq:QMIX_loss}
L(\theta) = \frac{1}{B} \sum_{i=1}^{B} H(R_{tot}^{i} - Q_{tot}(s, a, \theta))
\end{equation}

where $\theta$ are parameters of the agent network, the mixing network and the hyperparameter networks together. $H$ is Huber loss function defined in equation \ref{eq:huber_loss}. $B$ is the batch size and $R_{tot}$ is the global reward for the episode which is equal to  $(-1)\ \times$ (duration of the episode).


QMIX works better than other methods including \emph{Value Decomposition Networks} \cite{sunehag2017value}  that also satisfy Equation \ref{eq:consistency}, because it is able to represent
a larger family of monotonic functions. In addition, full factorisation
of the action-value function is not necessary to extract the decentralized policies.

\section{Experimental Results}
\label{sec:experiment}

In this section, several implementation details and experiment results are presented, and the simulation environment mentioned in Section \ref{sec:sim_env} is used. The experiments are divided into three parts. First, the model is trained on a $100 \times 100$ grid map with different combinations of cars and passengers and tested on the same map. In the second experiment, the performance on variable number of cars and passengers is evaluated, meaning numbers of cars and passengers is randomly assigned between episodes. Finally, a model trained on a $100 \times 100$ map is tested on different map sizes with different combinations of cars and passengers in order to test how the learnt strategies generalize to other map sizes. 

For each experiment, four different methods are used for comparing. The first one is Random, in which passengers are randomly assigned cars. The methods: greedy, IDQN, and QMIX are mentioned in section \ref{sec:method}. $P$ indicates the number of passengers in the experiment and $C$ stands for the number of cars. The values in the table are the average steps to drop off all passengers over $1,000$ episodes. A shorter episode time means better performance because it corresponds to all passengers being dropped off sooner.

Hyperparameters such as hidden size, learning rate, hidden size of the mixing network, batch size, epsilon decay, and number of episodes needed to be tuned. To keep results consistent, the hidden size, learning rate, batch size, epsilon decay, and number of episodes were kept consistent across both QMIX and IDQN. Both RL algorithms trained for $50000$ episodes with a learning rate of $.001$, a hidden size of $128$, a batch size of $128$, and an epsilon decay of $20,000$. The epsilon decay affects how quickly a network learns, and having this value of epsilon decay resulted in a good balance between exploration and exploitation.


\subsection{Fixed Number of Passengers and Cars}

In the first experiment conducted, the number of cars and passengers were fixed for all episodes for a single configuration, and the grid size was set to $100$ x $100$. Different combinations of numbers of cars and passengers were compared in this fixed configuration. QMIX outperformed all of the other methods for each of the configurations tested. When there is a large number of cars and passengers such as in $P=25,C=20$; QMIX is 18.9\% faster than greedy and 9.8\% faster than IDQN. When there are more passengers than cars, such as in $P=7,C=2$, QMIX was 8.7\% better than greedy and 2.7\% faster than IDQN. When there were more cars than passengers such as in $P=11,C=13$, QMIX performed 6.6\% faster than Greedy and 4.7\% faster than IDQN.

\begin{table}[]
\caption{Average episode duration for different configurations of passengers and cars}
\begin{tabular}{|c|c|c|c|c|c|c|}
\hline
\multicolumn{7}{|c|}{\textbf{100 x 100 Map Size}}     \\ \hline
\textbf{Method} & \textbf{P=7,C=2}  & \textbf{P=10,C=10} & \textbf{P=11,C=13} & \textbf{P=9,C=4}  & \textbf{P=10,C=2} & \textbf{P=25,C=20} \\ \hline
Random & 3386.248 & 2210.87   & 2089.87   & 2958.861 & 4644.59  & 2962.79   \\ \hline
Greedy & 3526.959 & 2208.55   & 2089.63   & 3072.806 & 4934.91  & 3173.54   \\ \hline
DQN    & 3306.884 & 2102.65   & 2046.65   & 2763.201 & 4847.97  & 2853.66   \\ \hline
QMIX (ours)     & \textbf{3218.542} & \textbf{2042.44}   & \textbf{1951.378}  & \textbf{2724.029} & \textbf{4357.72}  & \textbf{2573.24}   \\ \hline
\end{tabular}
 
\label{tab:diff_config}
\end{table}

\subsection{Variable Number of Cars and Passengers}
In the second experiment conducted, the number of cars and passengers was allowed to vary in between episodes. The average episode duration over 1000 testing episodes can be seen in Table \ref{tab:var_num_car_pass}. For each episode, the number of cars and passengers was sampled randomly from 1 to a predetermined maximum number of cars $C_{max}$  and passengers $P_{max}$. 

This is a more difficult and more realistic problem than a fixed number of cars and passengers since the strategy must adapt to more changing environmental factors including the number of agents. Nevertheless, QMIX was able to outperform the other methods in each of the configurations tested. For a $10\times10$ grid map with a maximum 10 cars and passengers, QMIX was 6.4\% better than Random, and 1.9\% better than IDQN. For the $500\times500$ grid map with a max of 20 cars and passengers, QMIX performed 7.6\% better than greedy and 4.8\% better than IDQN. 

\begin{table}[]
\caption{Average episode duration for a variable number of cars and passengers} 
\centering
\begin{tabular}{|c|c|c|}
\hline
 & \textbf{10 x 10 Map Size} & \textbf{500 x 500 Map Size} \\ \hline
\textbf{Method} & \textbf{Pmax= 10, Cmax= 10} & \textbf{Pmax= 20, Cmax= 20} \\ \hline
Random & 209.03 & 13700.14 \\ \hline
Greedy & 201.85 & 13871.07 \\ \hline
DQN & 199.43 & 13462.82 \\ \hline
QMIX(ours) & \textbf{195.66} & \textbf{12812.79} \\ \hline
\end{tabular}

\label{tab:var_num_car_pass}
\end{table}

\subsection{Generalization to Different Grid Sizes}
In the third experiment conducted, the models trained in the first experiment grid were tested on different grid sizes to see how well the strategies learned on a $100\times100$ grid generalize to different sizes. The results obtained can be seen in Table \ref{tab:var_grid_size}.

For nearly every configuration, QMIX generalized to the other sizes tested better than any other method. For example, for $P=7,C=2$ on a $10\times10$ grid, the generalized QMIX performed 9.2\% better than greedy and 2.3\% better than the generalized IDQN. Additionally, for $P=25,C=20$ on a $500\times500$ grid map, the generalized QMIX performed 20.3\% better than greedy and 12.7\% better than the generalized IDQN. Even in the cases where IDQN generalized better than QMIX, QMIX was still comparable and was never more than 2.5\% slower than IDQN. 

This result shows that QMIX can be trained on one map size and tested on another map size, either larger or smaller, and still perform fairly well. It was also able to generalize to other sizes better than IDQN.

\begin{table}[]
\caption{Training on a 100x100 grid and testing on other grid sizes} 
\begin{tabular}{|c|c|c|c|c|c|c|}
\toprule
\multicolumn{7}{|c|}{\textbf{10 x 10 Map Size}}   \\ \hline
\textbf{Method} & \textbf{P=7,C=2} & \textbf{P=10,C=10} & \textbf{P=11,C=13} & \textbf{P=9,C=4}  & \textbf{P=10,C=2} & \textbf{P=25,C=20} \\ \hline
Random & 337.3   & 215.64    & 208.77    & 287.57   & 449.38   & 291.62    \\ \hline
Greedy & 348.7   & 209.07    & 199.75    & 303.86   & 474.44   & 287.84    \\ \hline
DQN    & 323.9   & \textbf{201.28}    & 197.53    & \textbf{262.40}   & 454.27   & 285.62    \\ \hline
QMIX(ours)      & \textbf{316.5}   & 206.46    & \textbf{197.32}    & 265.10   & \textbf{417.91}   & \textbf{283.06}    \\ \hline
\multicolumn{7}{|c|}{\textbf{500 x 500 Map Size}} \\ \hline
\textbf{Method} & \textbf{P=7,C=2} & \textbf{P=10,C=10} & \textbf{P=11,C=13} & \textbf{P=9,C=4}  & \textbf{P=10,C=2} & \textbf{P=25,C=20} \\ \hline
Random & 17092.4 & 10860.21  & 10428.24  & 14715.82 & 23303.64 & 14820.33  \\ \hline
Greedy & 17251.2 & 10720.69  & 10905.83  & 15582.50 & 24491.44 & 16046.29  \\ \hline
DQN    & 16473.1 & 10139.36  & 9968.44   & 13571.88 & 23688.50 & 14649.64  \\ \hline
QMIX(ours)      & \textbf{16274.3} & \textbf{10120.57}  & \textbf{9835.75}   & \textbf{13548.92} & \textbf{21871.99} & \textbf{12784.23}  \\ \hline
\end{tabular}

\label{tab:var_grid_size}
\end{table}



\section{Conclusion}

In this paper, we propose a new MARL approach based on QMIX \cite{rashid:icml18} which allows us to perform centralized training with decentralized execution. We compare our proposed method against the Random, Greedy as well as the IDQN method through various experiments. QMIX was able to complete the rides faster than the other methods for each configuration on a $100\times100$ grid with a fixed number of cars and passengers. Additionally, QMIX performed better than the other methods for a varying number of cars and passengers. Finally, once trained with a fixed number of cars and passengers, QMIX was able to generalize to other map sizes better than the other methods for nearly every configuration. Future directions for extending this work include considering a more realistic model for the weights in our maps from real-world datasets and also having a modifiable grid map where connections can be removed or added dynamically.

\section*{Broader Impact}

Ridesharing services like Uber and Lyft as well as regular taxi companies could be benefited from this research by allowing them to complete more rides in less time. Customers could also be benefited by having shorter waiting times. On the other hand, taxi and ridesharing companies without the infrastructure for real-time car and passenger tracking or without the capabilities to run neural networks at scale could be placed at a competitive disadvantage. 

Failures in our system could result in sub-optimal car to passenger assignments and could be costly for ridesharing companies. Given the nature of the data collected by our algorithm and our objective of minimizing overall waiting time for the passengers our method does not leverage any biases in the data.

\begin{ack}
All authors contributed equally to the work. Everyone met together multiple times per week to work on the conception of the problem statement and implementation of the simulation environment. After implementing an initial environment, the group members split into two groups of two working in parallel on implementing the four methods in two different environment situations. When one of this environments trained more efficiently, Ting-Sheng Chu began making the environment compatible with RLPYT. Oscar de Lima and Hansal Shah worked on implementing the QMIX and IDQN algorithms. Brian Fogelson and Hansal Shah began to test the networks with different configurations and different hyperparameters to evaluate performance. Then the whole group began running tests plus finalizing the presentation and paper together. 
The entirity of this project was funded by the authors themselves.
\end{ack}

\bibliographystyle{unsrtnat}
\bibliography{bibfile} 

\end{document}